%% file: main.tex
\pdfoutput=1
\documentclass[11pt]{article}

\usepackage[final]{acl}
\usepackage{fancyhdr}
\fancypagestyle{myformat}{
    \fancyhf{} % 同时清空页眉和页脚的所有内容
    \fancyfoot[C]{\small \textit{Findings of the Association for Computational Linguistics: EMNLP 2025}, pages 10842--10854 \\
    November 4-9, 2025 \copyright 2025 Association for Computational Linguistics}
}
\usepackage{times}
\usepackage{latexsym}

\usepackage[T1]{fontenc}
\usepackage[utf8]{inputenc}

\usepackage{microtype}

\usepackage{inconsolata}

\usepackage{graphicx}

\usepackage{amsmath}
\usepackage{amssymb}
\usepackage{array}
\usepackage{multirow}
\usepackage{booktabs}
\usepackage{xcolor}

\usepackage[listings]{tcolorbox}
\tcbuselibrary{listings,theorems}
\definecolor{red1}{HTML}{FFCCC9}
\definecolor{green1}{HTML}{B5FFAA}
\usepackage{colortbl}
\usepackage{enumitem}
\usepackage[linesnumbered,ruled,vlined]{algorithm2e}

\usepackage[utf8]{inputenc}  % 处理 UTF-8 编码字符
\usepackage{graphicx} % 加载图形处理宏包
\usepackage{booktabs}  % 更美观地绘制表格线
\usepackage{multirow}  % 多行合并单元格
\usepackage{caption}   % 添加 caption 设置
\usepackage{subcaption}

\definecolor{purple}{HTML}{8d3a94}
\newtcbtheorem[]{exmp}{Prompt}%
{colback=purple!5,colframe=purple!80,fonttitle=\bfseries, left=.02in, right=.02in,bottom=.02in, top=.02in}{exmp}

\title{Self-Guided Function Calling in Large Language Models via Stepwise Experience Recall}

\author{
 \textbf{Sijia Cui\textsuperscript{1,2}},
 \textbf{Aiyao He\textsuperscript{1}},
 \textbf{Shuai Xu\textsuperscript{3}},
 \textbf{Hongming Zhang\textsuperscript{1}},
 \textbf{Yanna Wang\textsuperscript{1}$^\dagger$},
\\
 \textbf{Qingyang Zhang\textsuperscript{1}},
 \textbf{Yajing Wang\textsuperscript{4}},
 \textbf{Bo Xu\textsuperscript{1}$^\dagger$}
\\
 \textsuperscript{1}The Key Laboratory of Cognition and Decision Intelligence for Complex Systems, \\Institute of Automation, Chinese Academy of Sciences\\
 \textsuperscript{2}School of Artificial Intelligence, University of Chinese Academy of Sciences\\
 \textsuperscript{3}Nanjing University of Information Science \& Technology\\
 \textsuperscript{4} Institute of Computing Technology, Chinese Academy of Sciences
\\
 \small{
   \textbf{$^\dagger$Correspondence: wangyanna2013@ia.ac.cn, boxu@ia.ac.cn}
 }
}

\begin{document}
\maketitle
\thispagestyle{myformat}

\begin{abstract}
Function calling enables large language models (LLMs) to interact with external systems by leveraging tools and APIs. When faced with multi-step tool usage, LLMs still struggle with tool selection, parameter generation, and tool-chain planning. Existing methods typically rely on manually designing task-specific demonstrations, or retrieving from a curated library. These approaches demand substantial expert effort and prompt engineering becomes increasingly complex and inefficient as tool diversity and task difficulty scale. To address these challenges, we propose a self-guided method, \textbf{S}tepwise \textbf{E}xperienc\textbf{E} \textbf{R}ecall (SEER), which performs fine-grained, stepwise retrieval from a continually updated experience pool. Instead of relying on static or manually curated library, SEER incrementally augments the experience pool with past successful trajectories, enabling continuous expansion of the pool and improved model performance over time. Evaluated on the ToolQA benchmark, SEER achieves an average improvement of 6.1\% on easy and 4.7\% on hard questions. We further test SEER on $\tau$-bench, which includes two real-world domains. Powered by Qwen2.5-7B and Qwen2.5-72B models, SEER demonstrates substantial accuracy gains of 7.44\% and 23.38\%, respectively.
\end{abstract}

\section{Introduction}
Large language models (LLMs) demonstrated remarkable capabilities through pretraining on large-scale corpora~\citep{brown2020language, devlin-etal-2019-bert, touvron2023llama, achiam2023gpt, denison2024sycophancy, team2023gemini}. However, due to the inherent limitations of neural network architecture, LLMs are unable to interact directly with the real world—an issue that cannot be resolved simply by scaling up the training data or model size. 
Function calling\footnote{We use function calling and tool-use interchangeably.}~\cite{qu2025tool, qin2024tool, qin2024toolllm} serves as a fundamental mechanism that enables LLMs to interact with external systems.
By invoking external tools, LLMs can integrate up-to-date knowledge and execute real-world tasks, thereby expanding the boundaries of LLM-based AI agents and driving advancements across various domains~\cite{hao-etal-2025-large, theuma-shareghi-2024-equipping, zhong2023llm4eda, zhao2024expel}.

In-context learning~\citep{brown2020language} enhances LLM reasoning by embedding task-specific examples in prompts, enabling adaptation to new tasks without training.
However, this approach faces significant challenges in multi-step tool-use scenarios. Limited by the maximum token length, prompts cannot include examples that comprehensively cover all tools and problem types. Moreover, the relevance and complexity of the demonstrations directly impact the model's performance~\citep{zhao2021calibrate,min-etal-2022-rethinking, dong-etal-2024-survey}. 
This raises the critical question: How can we dynamically select examples tailored to the specific problem at hand, especially when the task involves multiple steps and complex tool interactions?

Existing methods~\citep{paranjape2023art,guan2025deeprag} typically rely on coarse-grained retrieval strategies, which fail to account for the nuanced relationships between tool usage patterns and user objectives in multi-step function calling. These approaches emphasize task similarity while overlooking the critical role of tool-chain alignment in achieving accurate and efficient outcomes.
Additionally, several approaches~\citep{zhao2024expel, xu-etal-2024-enhancing-tool} depend on manually curated or pre-collected task-specific demonstrations. However, this reliance not only limits scalability but also incurs substantial offline costs, making it inefficient when addressing a wide range of diverse tasks.

We introduce \textbf{S}tepwise \textbf{E}xperienc\textbf{E} \textbf{R}ecall (SEER)\footnote{\url{https://github.com/AI-Research-TeamX/SEER}}, a novel approach that enhances the multi-step tool-use capabilities of LLMs through fine-grained retrieval. It selects relevant trajectories by jointly considering task similarity, toolchain coverage, and intent alignment.
SEER incrementally expands the experience pool by incorporating successful task trajectories, improving model performance over time. 
For tasks without explicit success signals, we adopt an LLM-as-a-judge mechanism~\cite{li2024generation} to assess task completion. This enables continuous online updates to the pool, allowing SEER to adapt to new tasks and evolving user demands.
Our main contributions are:
\begin{itemize}[leftmargin=0.2in]
    \item We propose stepwise experience recall, retrieving relevant examples based on trajectory similarity, toolchain coverage, and user intent. This fine-grained retrieval effectively leverages experience from prior successful trajectories.
    \item We introduce online experience accumulation, dynamically adding successful multi-step tool invocation trajectories to the experience pool. This reduces reliance on manual annotations and enables the model to online self-improve.
    \item We conduct comprehensive evaluations on ToolQA and $\tau$-bench, and the results show that SEER outperforms existing methods. Meanwhile, the self-improvement results show clear and consistent performance gains over time, demonstrating the effectiveness of SEER and the potential for self-guided function calling.
    \item We perform extensive ablation studies involving different retrieval strategies and few-shot settings, aiming to highlight the contribution of each scoring component and impact of the number of demonstrations on performance.
\end{itemize}

\section{Related Work}

\subsection{Multi-step Function Calling}

Recent progress in tool-augmented LLMs has focused on either freezing or training approaches \cite{wang2024tools, huang2023metatool, yu2024steptool, goldie2025synthetic}.
Many studies exploit LLMs' in-context learning by prompting task descriptions and tool-use examples during inference \cite{lu2023chameleon,shen2023hugginggpt,hsieh2023tool,paranjape2023art,bai2024bayesian,zhangamulet,yang2025coarse,xu2025strategy,cui2025empowering}. For example, the ART framework \cite{paranjape2023art} retrieves similar multi-step reasoning and traces to guide models in generating intermediate steps and invoking functions.
Other methods, such as StepTool \cite{yu2024steptool} and SWiRL \cite{goldie2025synthetic}, treat tool use as a reinforcement learning problem \cite{sutton2018reinforcement,dong2020deep,zhang2020taxonomy}. StepTool applies step-wise reward shaping and policy gradients to improve decision-making based on tool success and task contribution. SWiRL generates synthetic multi-step tool-use data and uses step-wise RL with reward models to train without manual labels.
In contrast, we introduce an experience replay approach that enhances multi-step tool use without additional training, using fine-grained replay to improve performance efficiently.

\subsection{Self-improvement for LLM}

Accelerated advancements in LLMs have intensified data scarcity issues, highlighting data bottlenecks as a major research challenge \cite{villalobos2022will}. Self-improvement involving model-generated data such as feedback, instructions, and questions, has shown promise but often relies on heuristics and human validation for quality assurance \cite{bai2022constitutional,wang2022self}. Systems like ExpeL \cite{zhao2024expel} leverage past task experiences to enhance decision-making at inference, and recent work by \citep{tian2024toward} integrates Monte Carlo Tree Search (MCTS) \cite{kocsis2006bandit,zhang2020alphazero} with language models, creating annotation-free self-improvement loops. However, these methods typically rely on static offline datasets, significantly limiting adaptability in practical scenarios. To address this, we propose an online updating experience pool that continuously supports in-context learning, enhancing inference quality and real-world adaptability.

\section{Method}

\begin{figure*}[htbp]
    \centering
    \includegraphics[width=1.0\textwidth]{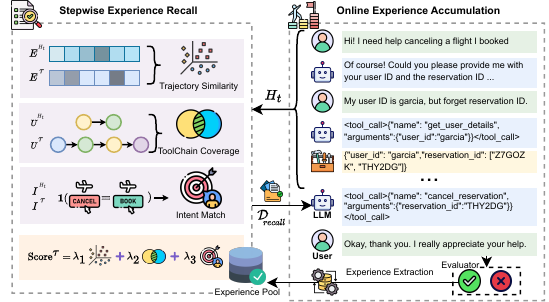}
    \caption{Overview of the SEER framework. The core component is the stepwise experience recall (left), which retrieves relevant trajectories from the experience pool based on the current interaction history $H_t$, and returns the top-$k$ examples $\mathcal{D}_{\text{recall}}$ to guide the LLM’s next decision. The continual experience accumulation mechanism (right) updates the experience pool by identifying successful trajectories using an evaluator.
    }
    \vspace{-0.1in}
    \label{fig:framework}
\end{figure*}

In this section, we present \textbf{S}tepwise \textbf{E}xperienc\textbf{E} \textbf{R}ecall (SEER), a novel approach that retrieves prior successful trajectories as in-context examples. 
SEER consists of three core components: trajectory experience extraction, stepwise experience recall, and continual experience accumulation, which together enable dynamic and efficient demonstration selection. 
The framework is illustrated in Figure~\ref{fig:framework}. We first present the notation and problem formulation, then detail each SEER component.

\subsection{Problem Formulation}

Building upon the formulation introduced in~\cite{zhao2024expel}, we consider an interactive task with tool-augmented large language model (LLM) agents. In this setting, an LLM-based agent is required to achieve a certain user-defined goal $g \in \mathcal{G}$ by interacting with a user and utilizing a set of external tools. The interaction unfolds over a finite horizon of $T$ steps, indexed by $i \in {0, \dots, T}$.

At each step $i$, the agent receives an observation $o_i \in \mathcal{O}$, where $\mathcal{O}$ denotes the joint observation space composed of both the user's input (or feedback) and the outputs returned by the tools. 
% Formally, we define the observation space as a Cartesian product: $\mathcal{O} = \mathcal{O}_{\text{user}} \times \mathcal{O}_{\text{tool}},$
% where $\mathcal{O}_\text{user}$ represents the space of user response, and $\mathcal{O}_\text{tool}$ denotes the space of tool output.
Formally, the observation space is the Cartesian product $\mathcal{O} = \mathcal{O}_{\text{user}} \times \mathcal{O}_{\text{tool}},$ where $\mathcal{O}_\text{user}$ is the user response space, and $\mathcal{O}_\text{tool}$ is the tool output space.

The agent maintains an interaction history $H_t = \{o_0, a_0, \dots, o_{t-1}, a_{t-1}, o_t\}$ up to the current time step $t$. Based on this history, the agent selects an action $a_t \in \mathcal{A}$, where $\mathcal{A}$ denotes the action space, including both natural language responses to the user and tool invocations. Once the action is executed, the agent receives a new observation $o_{t+1}$, which includes the user feedback or the tool result. It continues until the goal $g$ is achieved or a maximum interaction step $T$ is reached, finally yielding a complete trajectory $\tau = \{o_0, a_0, o_1, a_1, \dots\}$.

\subsection{Trajectory Experience Extraction}

Trajectory experience extraction transforms the interaction trajectory $\tau = \{o_0, a_0, o_1, a_1, \dots\}$ into a structured experience representation:

$$d^{\tau} = \langle E^{\tau}, E^q, I^{\tau}, U^{\tau} \rangle$$

Here, $E^{\tau}$ denotes the embedding of the interaction trajectory ${\tau}$, and $E^q$ is the embedding of the user's first query, the agent's first observation $o_0$, both generated using a pre-trained embedding model. $I$ represents the inferred user intent, selected from a predefined discrete intent set $\mathcal{I}$. Rather than relying on manual annotation, we employ the LLM itself to classify the user's goal based on the initial query. 
$U$ captures the sequence of tool invocations, which is modeled as a directed path $u_1 \rightarrow u_2 \rightarrow \dots \rightarrow u_n$, where each $u_i$ represents the $i$-th tool used within the overall trajectory.

\subsection{Stepwise Experience Recall}

Traditional retrieval methods primarily rely on the similarity between task instructions or user queries to select in-context examples.
However, this approach often fails to capture the dynamic nature of multi-turn interactions and multi-step tool invocation patterns. To address this limitation, we propose a multi-dimensional scoring strategy in SEER that considers three key aspects: trajectory similarity, toolchain coverage, and intent alignment.

First, we compute trajectory similarity by comparing the overall embeddings of interaction histories. This allows the recall mechanism to go beyond surface-level query matching and retrieve examples with similar structural progressions and decision patterns, providing richer contextual guidance. 
Then, we introduce a toolchain coverage score to account for similarities in tool usage sequences. Even when user goals appear superficially different, similar toolchains often reflect shared reasoning strategies or problem-solving procedures. By identifying examples with overlapping tool invocation patterns, SEER promotes the reuse of effective operational knowledge. 
Finally, we incorporate an intent match score based on inferred user intents. This helps the system better focus on the semantic core of the user’s goal. By aligning closely with user intent, SEER improves both the relevance and coherence of the retrieved demonstrations.

This multi-dimensional scoring enables SEER to recall more contextually appropriate and semantically aligned examples, thereby enhancing the agent's ability to generalize and perform effectively across diverse tool-usage tasks.
Specifically, given the current interaction history $H_t = \{o_0, a_0, o_1, a_1, \dots, o_t\}$ and a candidate trajectory $\tau' = \{o'_0, a'_0, o'_1, a'_1, \dots, o'_{t'}\}$ from the experience pool, we compute a relevance score between corresponding experience representations $d^{H_t}$ and $d^{\tau'}$. The score comprises the following components:

\begin{itemize}
    \item \textbf{Trajectory Similarity} ($s_1 \in [0, 1]$): Normalized cosine similarity between the embedding vectors:  
    $s_1 = (1 + \cos(E^{H_t}, E^{\tau'}))/{2}$

    \item \textbf{ToolChain Coverage} ($s_2 \in [0, 1]$): Proportion of tools in the current task that are also present in the candidate trajectory:
    \[
    s_2 = |U^{H_t} \cap U^{\tau'}| \,/\, |U^{H_t}|
    \]
    When calculating ToolChain Coverage, we ignore the directionality of $U$ and treat it as an unordered set of tools: $\{u_0, u_1, \dots\}$.

    \item \textbf{Intent Match} ($s_3 \in \{0, 1\}$): Whether the inferred user intents are identical:  
    $s_3 = \textbf{1}[I^{H_t} = I^{\tau'}]$, and $\textbf{1}$ is the indicator function.
\end{itemize}

The final relevance score is a weighted sum: $\text{Score}^{\tau} = \sum_{i=1}^{3} \lambda_i s_i$, 
where $\lambda_1, \lambda_2, \lambda_3$ are hyperparameters controlling the contribution of each component.
The top-$k$ demonstration $\mathcal{D}_{\text{recall}}$ with the highest relevance scores are selected as in-context exemplars to guide the LLM’s next decision $a_t$.
% This scoring mechanism enables more accurate, context-sensitive selection of exemplar trajectories, allowing the model to effectively reuse past experience in a few-shot manner for multi-step tool use.
The detailed pseudocode for the stepwise experience recall is presented in Algorithm~\ref{alg:stepwise_recall}.

\begin{algorithm}[htbp]
\caption{Stepwise Experience Recall}
\label{alg:stepwise_recall}
\KwIn{Current interaction history $H_t$, experience pool $\mathcal{D}$, $\lambda_1, \lambda_2, \lambda_3$}
\KwOut{Top-$k$ relevant trajectories $\mathcal{D}_{\text{recall}}$}

\ForEach{$(\tau', d^{\tau'}) \in \mathcal{D}$}{
    $d^{\tau'} = \langle E^{\tau'}, E^q, I^{\tau'}, U^{\tau'} \rangle$\;
    Extract $E^{H_t}, E^{q}, I^{H_t}, U^{H_t}$ from $H_t$\;
    $s_1 \gets (1 + \cos(E^{H_t}, E^{\tau'})) / 2$\;
    $s_2 \gets |U^{H_t} \cap U^{\tau'}| \,/\, |U^{H_t}|$\;
    $s_3 \gets \textbf{1}(I^{H_t} = I^{\tau'})$\;
    $\text{Score}^{\tau'} \gets \lambda_1 s_1 + \lambda_2 s_2 + \lambda_3 s_3$\;
    % Candidate $(\tau', \text{Score}^{\tau'})$\;
}
Sort all $(\tau', \text{Score}^{\tau'})$ pairs by $\text{Score}^{\tau'}$ in descending order\;
Select top-$k$ trajectories to form $\mathcal{D}_{\text{recall}}$\;
\KwRet{$\mathcal{D}_{\rm{recall}}$}
\end{algorithm}

\subsection{Continual Experience Accumulation}
Unlike prior methods that rely on static, offline datasets, SEER incrementally builds its trajectory pool during deployment. 
However, the online nature of this approach presents a challenge: the lack of explicit signals indicating task completion. Inspired by works such as \cite{li2024generation}, we leverage LLM's self-assessment capabilities to mitigate this issue. 
Specifically, after completing a task, the system performs self-evaluation by comparing its own output against the reference answer. The evaluator returns a binary judgment indicating whether task is a success or failure, determining whether the trajectory should be added to the experience pool. To ensure robustness, the evaluation logic is designed to tolerate minor discrepancies, such as formatting variations or slight numerical differences, and still regard them as successful outcomes.

By evaluating the correctness of the generated trajectory, we can determine whether to add the trajectory to the experience pool. This self-guided mechanism allows for continuous updates to the trajectory experience library, ensuring that the model remains adaptable and benefits from newly encountered cases without requiring extensive pre-collected data.
We show the continual experience accumulation process in Algorithm~\ref{alg:continual_experience_accumulation}. 
%方法的总结
We leave the implementation details of intent inference and LLM evaluator to the Appendix~\ref{apd:prompt_intent} and \ref{apd:prompt_evaluator}. 

\begin{algorithm}[htbp]
\caption{Experience Pool Update}
\label{alg:continual_experience_accumulation}
\KwIn{Current interaction history $H_t$, LLM evaluator $\mathcal{E}$, experience pool $\mathcal{D}$}
\KwOut{Updated experience pool}

\If{$\mathcal{E}.isSuccessful(H_t)$}{
    Extract $E^{\tau}, E^{q}, I^{\tau}, U^{\tau}$ from $H_t$\;
    Form experience tuple $d^{\tau} \gets \langle E^{\tau}, E^q, I^{\tau}, U^{\tau} \rangle$\;
    Insert $(\tau, d^{\tau})$ into experience pool $\mathcal{D} \gets \mathcal{D} \cup \{(\tau, d)\}$\;
}
\KwRet{$\mathcal{D}$}
\end{algorithm}

\section{Experiments}

We conducted extensive experiments to evaluate the performance of SEER on two benchmarks: ToolQA and $\tau$-bench. We first present the experimental setup, including benchmarks, evaluation metrics, and baseline methods. We then show the main results and conduct ablation studies to assess the contribution of core components in SEER. Finally, we highlight several insightful findings. In all result tables, the best performance is indicated in \textbf{bold}.
Our experiments are comprehensively designed to answer the following key questions:

\begin{itemize}[leftmargin=0.2in]
    \item How does SEER perform on both ToolQA and $\tau$-bench compared to existing baselines?
    \item Can SEER self-improve? Does its performance increase as the experience pool grows?
    \item How do different retrieval strategies impact the overall performance of SEER? In particular, does fine-grained retrieval yield better results?
    \item What is the effect of varying the top-$k$ value in SEER's stepwise experience recall mechanism?
\end{itemize}

%我们构建的实验主要为了探究几个问题：
%- SEER在ToolQA和$\tau$-bench上的整体表现如何？
%- SEER的自我改进能力如何？SEER的表现是否随着经验池的更新而提升？
%- SEER的不同检索方法对性能的影响如何？细粒度的检索是否能提升性能？
%- SEER的召回的top-k值对性能的影响如何？

\subsection{Experimental Setup}
We use Qwen2.5-72B-Instruct~\citep{qwen2025qwen25technicalreport} as the foundational LLM for both baseline methods and SEER. To ensure reproducibility and reduce randomness, we set model temperature to 0 across all experiments. For embedding-based retrieval, we adopt the bge-large-en-v1.5 model~\citep{xiao2024c}. Unless otherwise specified, the number of retrieved examples (top-$k$) is set to 4. The hyperparameters $\lambda_1, \lambda_2, \lambda_3$ are set to 1/3, 1/3, and 1/3, respectively. Maximum interaction steps $T=6$ for ToolQA and $T=30$ for $\tau$-bench. The maximum size of the experience pool is set to 1000. 
We initialize the demonstration pool with 8 same examples for ART, ExpeL, and SEER in ToolQA benchmark. In the $\tau$-bench setting, we initialize the demonstration pool with 2 examples and use GPT-4o as the user in the simulated environment.

\textbf{Benchmarks.} 
We primarily evaluated SEER on two challenging benchmarks designed for tool-augmented LLMs, shown in Table~\ref{tab:toolqa-statistics} and Table~\ref{tab:tau-bench-statistics}.
ToolQA~\citep{zhuang2023toolqa} spans 8 real-world domains—air transportation, financial data, commercial services, lodging platforms, social networks, academic publications, personal agendas, and numerical reasoning. The easy set comprises 800 questions across 55 templates, while the hard set includes 730 questions from 62 templates. ToolQA assesses LLMs' ability to reason across multi-step and use external tools effectively. $\tau$-bench~\citep{yao2025taubench} evaluates tool use in realistic, multi-turn tasks across airline (115 tasks, 15 tools) and retail (50 tasks, 13 tools) domains. Each task includes a user model and an LLM agent, simulating dynamic interactions and tool usage. Unlike static, single-turn question-answering settings, $\tau$-bench is specifically designed to assess LLM performance in dynamic, real-world scenarios, emphasizing multi-turn interaction, evolving user intent, and multi-step tool use.

\begin{table}[h]
    \centering
    \caption{An overview of the statistics of ToolQA.}
    \label{tab:toolqa-statistics}
    \resizebox{\linewidth}{!}{%
        \begin{tabular}{lllrrrrr}
        \toprule
        Domain & Data Type & Data Volume & \multicolumn{2}{c}{Easy Questions} & \multicolumn{2}{c}{Hard Questions} \\
        \cmidrule(lr){4-5} \cmidrule(lr){6-7}
        & & & Template & Count & Template & Count \\
        \midrule
        Flight & Structured DB & 4,078,318 & 10 & 100 & 10 & 100 \\
        Coffee & Structured DB & 5,746 & 8 & 100 & 13 & 130 \\
        Yelp & Structured DB & 150,346 & 11 & 100 & 10 & 100 \\
        Airbnb & Structured DB & 102,599 & 10 & 100 & 10 & 100 \\
        GSM8K & Professional & - & - & - & - & - \\
        DBLP & Graph DB & 553,320 & 10 & 100 & 10 & 100 \\
        SciREX & Text Corpus & 438 & 1 & 100 & 4 & 100 \\
        Agenda & Text Corpus & 10,000 & 5 & 100 & 5 & 100 \\
        \midrule
        Total & - & - & 55 & 800 & 62 & 730 \\
        \bottomrule
        \end{tabular}
    }
\end{table}

\begin{table}[h]
\centering
\caption{An overview of the statistics of $\tau$-bench.}
\label{tab:tau-bench-statistics}
\resizebox{\linewidth}{!}{%
    \begin{tabular}{llrr}
    \toprule
    Domain & Databases & Tools & Questions \\
    \midrule
    $\tau$-retail & 500 users, 50 products, 1,000 orders & 15 & 115 \\
    $\tau$-airline & 500 users, 300 flights, 2,000 reservations & 13 & 50 \\
    \midrule
    Total & - & - & 165 \\
    \bottomrule
    \end{tabular}
}
\end{table}

\input{table/main_result_toolqa}

\textbf{Evaluation Metrics.}
We use accuracy as the primary evaluation metric for assessing model performance on ToolQA. For a given question set \( Q_j \), the accuracy is defined as:
$Acc_j = \frac{1}{|Q_j|} \sum_{i=1}^{|Q_j|} \mathbf{1}[y_i = y'_i],$
where \( y_i \) is the ground-truth and \( y'_i \) is the predicted answer for the \( i \)-th question in $Q_j$.
To evaluate performance across multiple domains, we compute the average accuracy as:
$Acc^{\text{avg}} = \frac{1}{|D|} \sum_{j=1}^{|D|} Acc_j,$
where \( |D| \) denotes the number of domains in ToolQA, and \( \text{Accuracy}_j \) is the accuracy within the \( j \)-th domain.
In $\tau$-bench, $\text{pass\^{ }k}$ is a metric used to evaluate the probability that an LLM agent successfully completes the same task in all k independent dialogue trials.\footnote{To avoid ambiguity, we clarify that in this context, \(k\) refers to the number of trials. In all other parts of the paper, \(k\) denotes the number of demonstrations.} The metric is defined as:
$\text{pass\^{ }k} = \mathbb{E}_{\text{task}} \left[ \binom{c}{k} \mathbin{/} \binom{n}{k} \right],$
where \(c\) is the number of successful trials out of \(n\) total trials for a given task, and the expectation is taken over all tasks. 
Although $\text{pass\^{ }k}$ captures robustness over repeated attempts, we report $\text{pass\^{ }1}$ by default.
This simplifies the average reward (i.e., success rate) across tasks and serves as a standard baseline for evaluating the agent’s single-shot effectiveness.

\textbf{Baselines.}
To rigorously evaluate SEER, we compare it against five representative baselines: Chameleon~\cite{lu2023chameleon}, ReAct~\cite{yao2023react}, TUMS~\cite{he2025tums}, ART~\cite{paranjape2023art}, and ExpeL~\cite{zhao2024expel}. These methods cover a broad spectrum of strategies for tool-augmented LLMs, ranging from direct prompting to sophisticated reasoning and multi-step tool use.
For evaluation on $\tau$-bench, we include three closed-source and three open-source LLMs as reference models. Specifically, we test both the original versions of Qwen2.5-7B and Qwen2.5-72B, as well as their SEER-enhanced counterparts, to quantify the performance improvement introduced by our method.
Appendix~\ref{apd:baselines_models} shows the details of baselines.

\input{table/main_result_tau}

\subsection{Main Results}

Table~\ref{tab:main_result_toolqa} shows the results on the ToolQA benchmark. Our method, SEER, achieves consistent and significant improvements across multiple datasets. On easy questions, SEER outperforms the strongest baseline, ExpeL~\cite{zhao2024expel}, by 6.1\% in average accuracy. For hard questions, the improvement is 4.7\%.
Overall, SEER achieves an average accuracy of 67.9\% on easy sets and 31.1\% on hard sets. Specifically, SEER attains the best results on both easy and hard subsets of Yelp, Airbnb, DBLP, and Agenda. On Flight-Hard and Scirex-Hard, it shows a slight performance drop compared to the best baseline.
Notably, performance on Coffee-Easy declines more noticeably. We attribute this to suboptimal and cumbersome examples, which led to overthinking simple tasks. To address this, we incorporate reflection~\cite{shinn2023reflexion} to identify and correct errors. With this enhancement, SEER+Reflection achieves 69.0\% and 32.0\%. Detailed results are provided in Appendix~\ref{apd:more_results}.

The results on the $\tau$-bench benchmark are shown in Table~\ref{tab:main_result_tau}. Using the SEER method with the Qwen2.5-7B model, performance on the Airline task improves from 8.16\% to 20.41\%, and on the Retail task from 10.53\% to 13.16\%, yielding an overall average gain from 9.34\% to 16.78\%.
A similar trend is observed with the more powerful Qwen2.5-72B model, showing consistent performance improvements across both tasks. Notably, SEER equipped with Qwen2.5-72B achieves a final performance of 51.84\%, with a modest gap compared to the GPT-4o, which scores 54.76\%.

Overall, across eight datasets of varying difficulty and two real-world tasks, SEER consistently outperforms existing baseline methods, demonstrating the effectiveness and robustness of SEER in enhancing the multi-step and multi-turn function-calling capabilities of LLMs. To further investigate the nature of SEER’s self-guided mechanism and the impact of SEER components, we present detailed analyses from self-improvement experiments and ablation studies in the following sections.

\subsection{Self-Improvement}
\begin{figure}[h]
    \centering
    \includegraphics[width=1.0\linewidth]{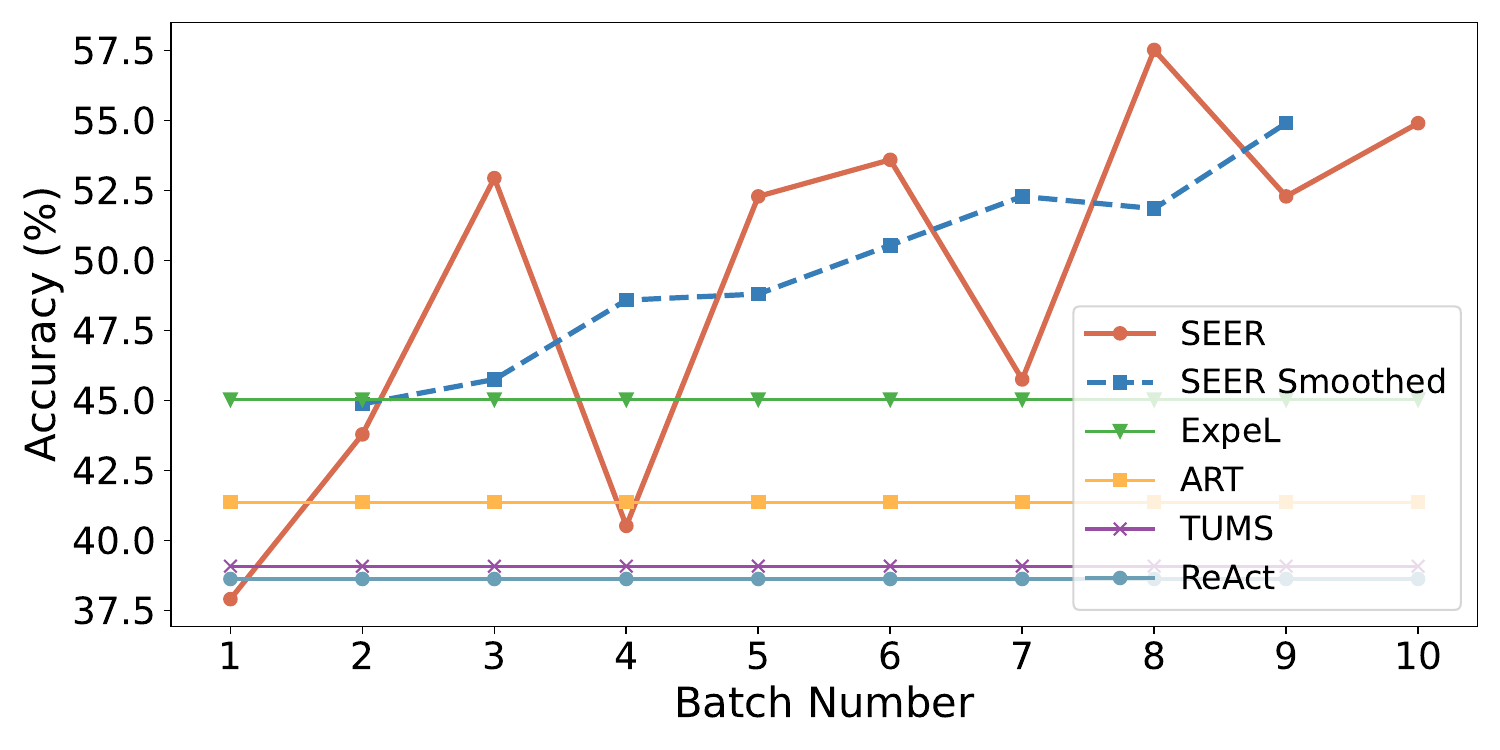}
    \caption{The self-improvement of SEER. The red solid line represents SEER's average accuracy per batch. The blue dashed line represents a 3-point moving average.}
    \label{fig:improvement}
\end{figure}

A key distinction between our method and existing approaches lies in how the demonstration pool is updated: instead of relying on a static, pre-collection pool, our method adopts an online self-guided mechanism. After each successful task completion, the corresponding trajectory is added to the demonstration pool, enabling continual refinement and enrichment over time.
To assess the self-improvement mechanism, we conduct a controlled experiment on the ToolQA benchmark. Specifically, we randomly shuffle all 1,530 questions and divide them into 10 batches, each containing 153 questions. For each batch, we perform offline evaluation and, upon completion, add all correctly answered instances into the demonstration pool for retrieval in the next batch.
We compute the batch accuracy for each batch: 
$Acc_j^{\text{batch}} = \frac{1}{|Q_{b_j}|} \sum_{i=1}^{|Q_{b_j}|} \mathbf{1}[y_i = y'_i],$ where $|Q_{b_j}|$ is the number of questions in the batch $b_j$, $y_i$ and $y'_i$ is the ground-truth and predicted answer for the $i$-th question in batch $b_j$. The 3-point moving average is used to smooth the results, which is defined as: $\hat{Acc}_j^{\text{batch}} = (Acc_{j-1}^{\text{batch}} + Acc_{j}^{\text{batch}} + Acc_{j+1}^{\text{batch}}) /\ {3}$.

We visualize the accuracy trend across successive batches in Figure~\ref{fig:improvement}, where a clear upward trajectory is observed. 
Initially, SEER's accuracy is relatively low, at 37.7\% at the first batch. However, as more batches are processed, SEER begins to effectively leverage the experience pool through stepwise recall. By the fifth batch, SEER surpasses baseline methods, reaching 52.3\% accuracy. This improvement continues, culminating in a batch accuracy of 54.9\% at the last batch. 
Overall, the consistent upward trend in performance underscores SEER’s capacity for self-improvement, validating the effectiveness of its self-guided learning strategy.
These highlight SEER’s potential for continual self-guidance and adaptation in real-world scenarios.

\input{table/few_shot.tex}

\subsection{Ablation Studies}
In addition to the main experiments described above, we have conducted ablation studies on the ToolQA benchmark to evaluate the effectiveness of different components in our method. 
Specifically, we experimented with various retrieval strategies and different numbers of demonstrations to reveal their contributions to overall performance.

\textbf{Retrieval Strategies.}
The retrieval score in SEER consists of three components: trajectory embedding similarity $s_1$, toolchain coverage $s_2$, and intent match score $s_3$. We compare different retrieval configurations, including SEER (w/o $s_2$) and SEER (w/o $s_3$), to analyze the impact of $s_2$ and $s_3$ on retrieval performance.
For $s_1$, we also explore a query-based variant method, SEER (query-based).
Specifically, instead of computing similarity over the entire trajectory, we compute $s_1 = (1 + \cos(E^{q}, E^{q'}))/{2}$, where $q$ is the first observation $o_0$, i.e., the user's initial query. This allows us to isolate and examine the effectiveness of query-level semantic similarity.

\begin{figure}
    \centering
    \includegraphics[width=1.0\linewidth]{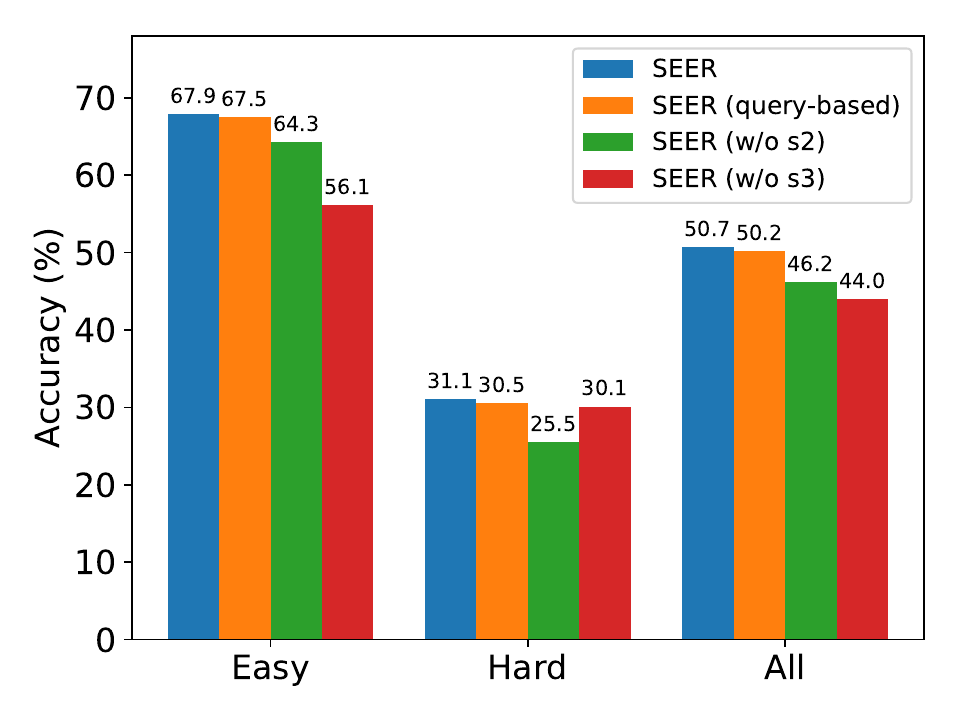}
    \caption{Accuracy of SEER and its ablated variants, showing the impact of each retrieval component.}
    % \vspace{-0.1in}
    \label{fig:retrieval_method}
\end{figure}

As shown in Figure~\ref{fig:retrieval_method}, all three retrieval variants result in performance degradation compared to the full SEER method. The performance drop in SEER (query-based) is relatively modest. This is expected, as the ToolQA benchmark adopts a one-turn multi-step setting where the user's query does not change across turns, making full-trajectory and query-only representations less divergent.
However, a more pronounced drop is observed in SEER (w/o $s_3$) when applied to easy questions. This indicates that retrieving exemplars with aligned user intent is particularly beneficial for tasks involving straightforward reasoning and tool usage.
In contrast, SEER (w/o $s_2$) shows the most significant decline on hard questions. It suggests that the toolchain coverage score is crucial for complex tasks, where the model must navigate intricate dependencies and interactions among multiple tools. 

In short, these results validate the effectiveness of our multidimensional retrieval scoring mechanism. 
% Specifically, the $s_2$ and $s_3$ components contribute significantly to performance on complex and simple tasks, respectively.
The $s_3$ component plays a crucial role in simpler tasks that benefit from retrieved intent-aligned exemplars. In contrast, the $s_2$ contributes significantly to performance on complex tasks, where multi-step tool use and dependencies are common.

\textbf{Demonstration Number.}
In this section, we analyze the impact of Top-$k$ on the performance of SEER. We vary the few-shot number from 0 to 8 and evaluate the model's performance on both easy and hard questions. 
From the Table~\ref{tab:few_shot_combined_bold}, we can observe a trend of performance improvement with increasing demonstration numbers, followed by a decline after reaching a peak. Specifically, on the easy question set, the model achieves its best performance at 4 demonstrations, with an accuracy of 67.88\%. On the hard question set, the model performs best at 4 demonstrations, achieving an accuracy of 31.51\%. We also test on query-based SEER, which shows a similar trend, with the best performance at $k=4$ for easy questions and $k=6$ for hard questions. This indicates that the model benefits from a moderate number of demonstrations, which provide sufficient context without overwhelming it with excessive information. 
Results on $\tau$-bench across different $k$ values exhibit a similar trend, as detailed in Appendix~\ref{apd:demonstration_number}.

% These findings not only demonstrate the effectiveness of adaptive prompting and the importance of updating the example library, but also suggest that more examples do not necessarily lead to better performance. Instead, there needs to be a balance between the quantity and quality of examples. The static method performs relatively poorly, especially on complex database query tasks. This is primarily because the examples included in the static prompt have low similarity to the target questions and contain irrelevant database queries and tool usages. These unrelated examples fail to provide useful guidance for answering the questions and instead interfere with the LLM's reasoning process, resulting in lower accuracy.

\section{Conclusion}

In this paper, we proposed SEER, a self-guided approach for tool-augmented LLMs. SEER introduces a stepwise experience recall mechanism to retrieve relevant past experiences and guide multi-step tool usage. It continually updates its experience pool with successful trajectories, enabling iterative self-improvement during deployment. 
% We conducted extensive experiments on multi-step and multi-turn benchmarks, demonstrating that SEER outperforms existing methods. The ablation studies further validate the effectiveness of SEER's components, including fine-grained retrieval and experience accumulation.
By conducting extensive experiments, we demonstrated that SEER significantly outperforms existing methods on both multi-step and multi-turn benchmarks. We also validated the self-improvement capability of SEER through intermediate batch accuracy evaluations. Additionally, we performed ablation studies to assess the contributions of SEER's components. 
Overall, the experiments show some key findings: (1) SEER is effective in improving the performance of LLMs on complex function calling tasks; (2) SEER's self-guided mechanism enables continual self-improvement; (3) The multi-dimensional retrieval strategy enhances the model's ability across different task scenarios; (4) The number of demonstrations plays a crucial role, with a moderate number yielding the best results.

\clearpage
\section*{Limitations}
While SEER demonstrates strong performance improvements in multi-step tool usage, several limitations remain. The diversity of the experience memory is inherently constrained by the capabilities of the underlying LLM. For complex or edge-case queries that require advanced reasoning beyond what can be addressed through in-context learning, SEER's self-guided mechanism may encounter performance bottlenecks. SEER uses a fixed retrieval weighting scheme across all tasks, which may not be optimal for heterogeneous domains or task types. Dynamic adaptation of retrieval strategies—such as learning task-aware weighting or incorporating uncertainty estimates—could further enhance SEER’s generalization and robustness.

\section*{Acknowledgments}
This work is supported by the National Key R\&D Program of China
(No.2022ZD0116405).
\bibliography{custom}

\clearpage
\appendix
\section*{Appendix}
\section{Prompt Template}
% \subsection{SEER Prompt}\label{apd:seer_prompt}

\subsection{Intent Inference Prompt}\label{apd:prompt_intent}

The prompt construction is shown in Figure \ref{fig:prompt_intent}.

\begin{figure}[t]
    \centering
    \input{prompt/intent}
    \caption{The illustration of the intent recognizer.}
    \label{fig:prompt_intent}
\end{figure}

\subsection{Trajectory Evaluation Prompt}\label{apd:prompt_evaluator}

The prompt construction is shown in Figure \ref{fig:evaluator}.

\begin{figure}[t]
    \centering
    \input{prompt/evaluator}
    \caption{The illustration of the evaluator.}
    \label{fig:evaluator}
\end{figure}

\input{table/apd_main_result_toolqa}

\section{Baselines}\label{apd:baselines_models}
We compare SEER with the following baselines, which are widely used in the field of tool-augmented LLMs. The details of these methods are as follows:

\begin{itemize}[leftmargin=*]
    \item \textbf{Direct, CoT~\cite{wei2022chain}}: Follow two baselines setting in~\cite{zhuang2023toolqa}, where the LLM is directly prompted with the user question and generate a response without awareness of tool invocation, to demonstrate the limitations of LLMs without tool assistance.
    \item \textbf{CoT-tool}: CoT with tool invocation. This method is similar to the original CoT but includes a tool interface for LLMs to invoke external tools.
    \item \textbf{Chameleon}: Chameleon~\cite{lu2023chameleon} is a plug-and-play compositional reasoning framework where the LLM acts as a controller to plan and execute tool chain. Each tool operates as an independent module, allowing flexible combinations and extensions for complex tasks.
    \item \textbf{ReAct}: ReAct~\cite{yao2023react} enables LLMs to alternately generate reasoning traces and task-specific actions, forming an iterative Observation-Thought-Action cycle. Compared to Chameleon, ReAct receives immediate feedback from tool executions, facilitating more adaptive decision-making.
    \item \textbf{TUMS}: TUMS~\cite{he2025tums} is a framework that enhances LLM' tool-use abilities by introducing fine-grained, parameter-level processing, enabling more accurate and reliable tool execution.
    \item \textbf{ART}: ART~\cite{paranjape2023art} is a multi-step reasoning and tool-use framework that retrieves similar task trajectories to guide the LLM in generating intermediate steps and invoking functions.
    \item \textbf{ExpeL}: ExpeL~\cite{zhao2024expel} is a self-improvement framework that enables LLMs to learn from past experiences and improve their performance over time.
\end{itemize}

In $\tau$-bench, We initially assessed the performance of three close source models (claude-3-5-sonnet-20241022, gpt-4o-2024-11-20, and gpt-4o-mini-2024-07-18) and three open-source models (Qwen2.5-7B-Instruct, Qwen2.5-72B-Instruct, and DeepSeek-V3-0324) on the Tau-Bench benchmark. Subsequently, we evaluated the improvements achieved by applying the proposed SEER method to Qwen2.5 models on the same benchmark.

\section{Full Main Results}\label{apd:more_results}
The full results on ToolQA are shown in Table~\ref{tab:apd_main_result_toolqa}.

\section{Demonstration Number}\label{apd:demonstration_number}

\begin{figure}
    \centering
    \includegraphics[width=1\linewidth]{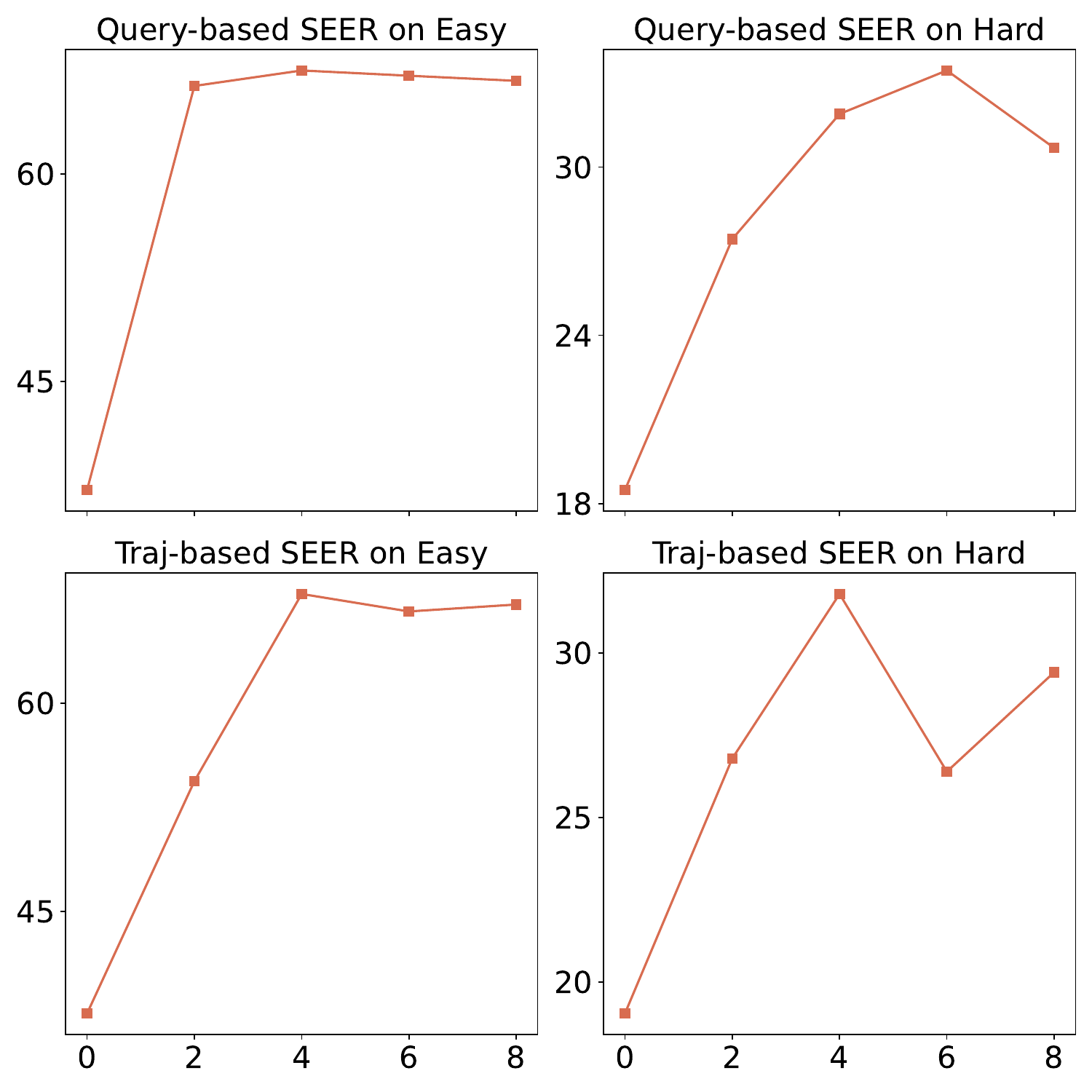}
    \caption{The accuracy of SEER with different few-shot numbers. The top plot shows query-based SEER, while the bottom plot shows SEER. The x-axis represents the number of few-shot demonstrations, and the y-axis represents the accuracy.}
    \label{fig:all_num_demos}
\end{figure}

The full ablation study of the number of demonstrations is shown in Figure~\ref{fig:all_num_demos}. The results show that the query-based SEER performs best with 4 demonstrations, achieving an accuracy of 67.5\% on easy questions and 33.56\% on hard questions. The performance declines when the number of demonstrations exceeds 4, indicating that too many examples can overwhelm the model and lead to confusion.

\section{Benchmark Details}\label{apd:benchmark}

\textbf{ToolQA.} 
The dataset is divided into two parts: simple and complex questions. The simple question set contains 55 templates, while the complex question set contains 62 templates. Each template is designed to cover a specific domain and includes a set of questions that can be answered using the tools provided in the benchmark. The dataset is designed to test the ability of LLMs to reason about and use external tools effectively. 
This large amount of data is mainly stored in the form of databases, graphs, and textual corpora, which greatly tests LLM's understanding and flexible application ability of the given tools. 

\textbf{$\tau$-bench.}
The dataset is divided into two parts: $\tau$-retail and $\tau$-airline. $\tau$-retail contains 115 questions, while $\tau$-airline contains 50 questions. 
In contrast to one-turn multi-step ToolQA, $\tau$-bench focuses on evaluating LLMs in real-world tasks through multi-turn user-agent interactions.

\end{document}

%% file: table/main_result_toolqa.tex
\begin{table*}[h]
\centering
\caption{Main results on the ToolQA Benchmark. The results are reported in terms of Accuracy (\%). The best results are highlighted in \textbf{bold}. Our method SEER achieves the best performance on average accuracy across all tasks, outperforming the second-best method ExpeL by 6.1\% and 4.7\% on \textbf{E}asy and \textbf{H}ard tasks, respectively.}
\label{tab:main_result_toolqa}
\resizebox{\textwidth}{!}{
\begin{tabular}{lcccccccccccccccccc}
\toprule
\multicolumn{1}{c}{\multirow{2}{*}{\textbf{Method}}} & \multicolumn{2}{c}{\textbf{Flight}} & \multicolumn{2}{c}{\textbf{Coffee}} & \multicolumn{2}{c}{\textbf{Yelp}} & \multicolumn{2}{c}{\textbf{Airbnb}} & \multicolumn{2}{c}{\textbf{DBLP}} & \multicolumn{2}{c}{\textbf{SciREX}} & \multicolumn{2}{c}{\textbf{Agenda}} & \textbf{GSM8K} & \multicolumn{2}{c}{\textbf{Avg.}} \\ \cmidrule(l){2-18} 
\multicolumn{1}{c}{} & \textbf{E} & \textbf{H} & \textbf{E} & \textbf{H} & \textbf{E} & \textbf{H} & \textbf{E} & \textbf{H} & \textbf{E} & \textbf{H} & \textbf{E} & \textbf{H} & \textbf{E} & \textbf{H} & \textbf{E} & \textbf{E} & \textbf{H} \\
\midrule
\textbf{Chameleon} & 37.0 & 2.0 & 61.0 & 2.3 & 60.0 & 13.0 & 12.0 & 4.0 & 25.0 & 17.0 & \textbf{6.0} & 19.0 & 57.0 & 0.0 & 17.0 & 34.5 & 8.2 \\
\textbf{CoT} & 37.0 & 22.0 & 79.0 & 13.1 & 43.0 & 52.0 & 79.0 & 19.0 & 0.0 & 1.0 & 3.0 & 22.0 & 54.0 & 0.0 & 2.0 & 37.1 & 18.4 \\
\textbf{ReAct} & 67.0 & 6.0 & 94.0 & 25.4 & 70.0 & 17.0 & 86.0 & 17.0 & 27.0 & 22.0 & \textbf{6.0} & 18.0 & 61.0 & 0.0 & 64.0 & 59.5 & 15.1 \\
\textbf{TUMS} & 64.0 & 9.0 & 93.0 & 22.3 & 73.0 & 15.0 & 91.0 & 11.0 & 34.0 & 30.0 & 4.0 & 14.0 & 59.0 & 0.0 & 72.0 & 61.3 & 14.5 \\
\textbf{ART} & 73.0 & 20.0 & \textbf{99.0} & 33.1 & 81.0 & 30.0 & \textbf{94.0} & 25.0 & 33.0 & 33.0 & 3.0 & \textbf{25.0} & 1.0 & 0.0 & 73.0 & 57.1 & 23.7 \\
\textbf{ExpeL} & 81.0 & \textbf{29.0} & 92.0 & 34.6 & 77.0 & 40.0 & 88.0 & 32.0 & \textbf{37.0} & 29.0 & 5.0 & 19.0 & 57.0 & 1.0 & 57.0 & 61.8 & 26.4 \\
\midrule
\textbf{SEER (Ours)} & \textbf{82.0} & 25.0 & 87.0 & \textbf{41.5} & \textbf{90.0} & \textbf{58.0} & \textbf{94.0} & \textbf{33.0} & \textbf{37.0} & \textbf{35.0} & \textbf{6.0} & 23.0 & \textbf{68.0} & \textbf{2.0} & \textbf{79.0} & \textbf{67.9} & \textbf{31.1} \\ \bottomrule
\end{tabular}}
\end{table*}

%% file: table/main_result_tau.tex
\begin{table}[!ht]
\centering
\caption{Main results on $\tau$-bench. With the integration of SEER, the performance of two open-source models is significantly enhanced. SEER (72B) achieves 51.84\%, approaching the performance of GPT-4o at 54.76\%.}
\label{tab:main_result_tau}
\resizebox{1.0\linewidth}{!}{
\begin{tabular}{@{}lccc@{}}
\toprule
\multicolumn{1}{c}{\textbf{Methods}} & \textbf{Airline} & \textbf{Retail} & \textbf{Avg.} \\ \midrule
\multicolumn{1}{c}{\textit{\textbf{\underline{Close source models}}}} &       &       &       \\
Claude 3.5 Sonnet                  & \textbf{48.98}   & \textbf{70.18}  & \textbf{59.58}   \\
GPT-4o                                        & 42.86 & 66.67 & 54.76 \\
GPT-4o-mini                                   & 22.45 & 47.37 & 34.91 \\ \midrule
\multicolumn{1}{c}{\textit{\textbf{\underline{Open source models}}}}  &       &       &       \\
DeepSeek-v3                                         & 46.94 & 67.54 & 57.24 \\
Qwen2.5-72B-Instruct                                             & 30.61 & 26.32 & 28.46 \\
Qwen2.5-7B-Instruct                                              & 8.16  & 10.53 & 9.34  \\ \midrule
\multicolumn{1}{c}{\textit{\textbf{\underline{Ours}}}}               &       &       &       \\
SEER(7B)                                                 & 20.41 & 13.16 & 16.78 \\
SEER(72B)                                                & 38.78 & 64.91 & 51.84 \\ \bottomrule
\end{tabular}}
\end{table}

%% file: table/few_shot.tex
\begin{table*}[!ht]
    \centering
    \caption{\enspace Performance on easy and hard questions under different top-$k$ few-shot settings. The results exhibit a trend of increasing performance followed by a decline as $k$ increases. SEER(Q) denotes SEER (\textbf{Q}uery-based).}
    \resizebox{\linewidth}{!}{
    \begin{tabular}{@{}cc|cc|cc|cc|cc|cc|cc|cc|cc@{}}
        \toprule
        \textbf{Method} & \textbf{Top-k} & \multicolumn{2}{c|}{\textbf{Flight}} & \multicolumn{2}{c|}{\textbf{Coffee}} & \multicolumn{2}{c|}{\textbf{Yelp}} & \multicolumn{2}{c|}{\textbf{Airbnb}} & \multicolumn{2}{c|}{\textbf{DBLP}} & \multicolumn{2}{c|}{\textbf{SciREX}} & \multicolumn{2}{c|}{\textbf{Agenda}} & \multicolumn{2}{c}{\textbf{Avg.}} \\
        \cmidrule{3-18}
        & & \textbf{E} & \textbf{H} & \textbf{E} & \textbf{H} & \textbf{E} & \textbf{H} & \textbf{E} & \textbf{H} & \textbf{E} & \textbf{H} & \textbf{E} & \textbf{H} & \textbf{E} & \textbf{H} & \textbf{E} & \textbf{H} \\ 
        \midrule

        \multirow{5}{*}{\textbf{SEER(Q)}} 
        & 0 & 37.0 & 22.0 & 79.0 & 13.08 & 43.0 & 52.0 & 79.0 & 19.0 & 0.0 & 1.0 & 3.0 & 22.0 & 54.0 & 2.0 & 37.13 & 18.49 \\
        & 2 & 81.0 & 18.0 & 95.0 & 40.0 & 80.0 & 60.0 & 94.0 & 21.0 & 33.0 & 27.0 & \textbf{9.0} & 22.0 & 65.0 & 4.0 & 66.38 & 27.95 \\
        & 4 & 79.0 & \textbf{39.0} & \textbf{97.0} & 39.23 & 80.0 & 61.0 & \textbf{95.0} & 24.0 & 33.0 & 26.0 & 4.0 & \textbf{24.0} & \textbf{69.0} & 5.0 & \textbf{67.5} & 31.51 \\
        & 6 & \textbf{87.0} & \textbf{39.0} & 96.0 & 43.08 & \textbf{82.0} & \textbf{64.0} & 90.0 & \textbf{28.0} & 34.0 & \textbf{32.0} & 5.0 & 20.0 & 66.0 & \textbf{6.0} & 67.13 & \textbf{33.56} \\
        & 8 & 86.0 & 24.0 & 90.0 & \textbf{50.77} & 80.0 & 62.0 & 89.0 & 24.0 & \textbf{36.0} & 30.0 & 8.0 & 21.0 & 66.0 & 1.0 & 66.75 & 31.23 \\
        \midrule

        \multirow{5}{*}{\textbf{SEER}} 
        & 0 & 37.0 & 21.0 & 80.0 & 16.92 & 43.0 & 53.0 & 81.0 & 19.0 & 0.0 & 1.0 & 3.0 & 21.0 & 55.0 & \textbf{2.0} & 37.63 & 19.04 \\
        & 2 & 73.0 & 20.0 & 87.0 & \textbf{41.54} & 81.0 & 52.0 & 92.0 & 30.0 & 31.0 & 23.0 & \textbf{6.0} & 20.0 & 64.0 & 0.0 & 58.88 & 27.26 \\
        & 4 & \textbf{82.0} & \textbf{25.0} & 87.0 & \textbf{41.54} & \textbf{90.0} & \textbf{58.0} & \textbf{94.0} & 33.0 & \textbf{37.0} & 35.0 & \textbf{6.0} & \textbf{23.0} & \textbf{68.0} & \textbf{2.0} & \textbf{67.88} & \textbf{31.51} \\
        & 6 & 70.0 & 17.0 & \textbf{98.0} & 40.77 & 85.0 & 50.0 & 92.0 & 28.0 & 33.0 & 27.0 & 5.0 & \textbf{23.0} & 65.0 & 0.0 & 65.88 & 27.12 \\
        & 8 & 81.0 & 21.0 & 95.0 & 33.85 & 78.0 & 56.0 & \textbf{94.0} & \textbf{36.0} & 34.0 & \textbf{40.0} & 5.0 & \textbf{23.0} & \textbf{68.0} & 0.0 & 67.13 & 30.14 \\
        \bottomrule
    \end{tabular}
    }
    \label{tab:few_shot_combined_bold}
\end{table*}

%% file: prompt/intent.tex
\begin{exmp}{Prompt of Intent Recognizer}{}
    \scriptsize
    You are an expert in \textbf{intent recognition}. Given a user question (from 8 distinct intent categories), your goal is to extract key information from the question and determine the most likely intent. \\
    
    You should think step by step and conclude your answer with an intent result in the format: \textbf{[INTENT]}, where \texttt{INTENT} is the name of the predicted intent category. \\[0.5em]
    
    \textbf{Below are descriptions of all 8 intent categories:} \\
    \{intent\_categories\} \\
    
    \textbf{Instructions:} Your response must follow this format:
    Question: This is the question. \\
    Answer: These are your thoughts. [INTENT] \\
    
    \textbf{Here are some examples:} \\
    \{examples\} \\
    
    \textbf{Now, please infer user intent:} \\
    Question: \texttt{\{question\}}
\end{exmp}

%% file: prompt/evaluator.tex
\begin{exmp}{Prompt of Evaluator}{}
    \scriptsize
    Please act as an evaluator to determine whether the model's response matches or includes the correct answer. 
    I will provide both the correct answer and the model's response. \\
    Please reply with either \textbf{[Match]} or \textbf{[No Match]}, and briefly explain the reasoning behind your judgment. \\[0.5em]

    \textbf{Examples:} \\[0.3em]
    \textbf{Question:} \texttt{What time is the meeting?} \\
    \textbf{Correct Answer:} \texttt{3:00PM} \\
    \textbf{Model Response:} \texttt{The meeting is scheduled for 15:00.} \\
    \textbf{Evaluation:} [Match] \\[0.7em]
    
    \textbf{Question:} \texttt{What is Mike's total cost?} \\
    \textbf{Correct Answer:} \texttt{9} \\
    \textbf{Model Response:} \texttt{The total cost of Mike is 9.001} \\
    \textbf{Evaluation:} [Match] \\[0.7em]
    
    \textbf{Question:} \texttt{When is he scheduled to attend the meeting?} \\
    \textbf{Correct Answer:} \texttt{01/12} \\
    \textbf{Model Response:} \texttt{He will attend this meeting on the morning of January 12th.} \\
    \textbf{Evaluation:} [Match] \\[0.7em]
    
    \textbf{Question:} \texttt{What is the price?} \\
    \textbf{Correct Answer:} \texttt{\$9374} \\
    \textbf{Model Response:} \texttt{None} \\
    \textbf{Evaluation:} [No Match] \quad (Model Response is None) \\[1em]
    
    \textbf{Now evaluate:} \\
    \textbf{Question:} \texttt{question} \\
    \textbf{Correct Answer:} \texttt{gt\_answer} \\
    \textbf{Model Response:} \texttt{llm\_answer}
\end{exmp}

%% file: table/apd_main_result_toolqa.tex
\begin{table*}[h]
\centering
\caption{Full main results on the ToolQA benchmark.}
\label{tab:apd_main_result_toolqa}
\resizebox{\textwidth}{!}{
\begin{tabular}{@{}lccccccccccccccccc@{}}
\toprule
\multicolumn{1}{c}{\multirow{2}{*}{\textbf{Method}}} & \multicolumn{2}{c}{\textbf{Flights}} & \multicolumn{2}{c}{\textbf{Coffee}} & \multicolumn{2}{c}{\textbf{Yelp}} & \multicolumn{2}{c}{\textbf{Airbnb}} & \multicolumn{2}{c}{\textbf{Dblp}} & \multicolumn{2}{c}{\textbf{Scirex}} & \multicolumn{2}{c}{\textbf{Agenda}} & \textbf{GSM8K} & \multicolumn{2}{c}{\textbf{Avg}} \\ \cmidrule(l){2-18} 
\multicolumn{1}{c}{} & easy & hard & easy & hard & easy & hard & easy & hard & easy & hard & easy & hard & easy & hard & easy & easy & hard \\ \midrule
\textbf{Direct} & 0.0 & 0.0 & 0.0 & 0.0 & 0.0 & 0.0 & 0.0 & 0.0 & 0.0 & 0.0 & 0.0 & 0.0 & 0.0 & 0.0 & 50.0 & 6.3 & 0.0 \\
\textbf{CoT-noTool} & 0.0 & 0.0 & 0.0 & 0.0 & 0.0 & 0.0 & 0.0 & 0.0 & 0.0 & 0.0 & 0.0 & 0.0 & 0.0 & 0.0 & 63.0 & 7.9 & 0.0 \\
\textbf{Chameleon} & 37.0 & 2.0 & 61.0 & 2.3 & 60.0 & 13.0 & 12.0 & 4.0 & 25.0 & 17.0 & 6.0 & 19.0 & 57.0 & 0.0 & 17.0 & 34.5 & 8.2 \\
\textbf{CoT} & 37.0 & 22.0 & 79.0 & 13.1 & 43.0 & 52.0 & 79.0 & 19.0 & 0.0 & 1.0 & 3.0 & 22.0 & 54.0 & 0.0 & 2.0 & 37.1 & 18.4 \\
\textbf{ReAct} & 67.0 & 6.0 & 94.0 & 25.4 & 70.0 & 17.0 & 86.0 & 17.0 & 27.0 & 22.0 & 6.0 & 18.0 & 61.0 & 0.0 & 64.0 & 59.5 & 15.1 \\
\textbf{TUMS} & 64.0 & 9.0 & 93.0 & 22.3 & 73.0 & 15.0 & 91.0 & 11.0 & 34.0 & 30.0 & 4.0 & 14.0 & 59.0 & 0.0 & 72.0 & 61.3 & 14.5 \\
\textbf{ART} & 73.0 & 20.0 & \textbf{99.0} & 33.1 & 81.0 & 30.0 & \textbf{94.0} & 25.0 & 33.0 & 33.0 & 3.0 & \textbf{25.0} & 1.0 & 0.0 & 73.0 & 57.1 & 23.7 \\
\textbf{ExpeL} & 81.0 & 29.0 & 92.0 & 34.6 & 77.0 & 40.0 & 88.0 & 32.0 & \textbf{37.0} & 29.0 & 5.0 & 19.0 & 57.0 & 1.0 & 57.0 & 61.8 & 26.4 \\
\textbf{ExpeL (8-shot)} & 84.0 & 27.0 & 88.0 & 30.8 & 77.0 & 35.0 & 93.0 & 24.0 & 35.0 & 28.0 & 5.0 & 21.0 & 65.0 & \textbf{2.0} & 70.0 & 64.6 & 24.0 \\ \midrule
\textbf{SEER (Ours)} & 82.0 & 25.0 & 87.0 & 41.5 & \textbf{90.0} & 58.0 & \textbf{94.0} & \textbf{33.0} & \textbf{37.0} & \textbf{35.0} & 6.0 & 23.0 & \textbf{68.0} & \textbf{2.0} & \textbf{79.0} & 67.9 & 31.1 \\
\textbf{SEER + Reflection} & \textbf{88.0} & \textbf{31.0} & \textbf{99.0} & \textbf{52.3} & 85.0 & \textbf{63.0} & \textbf{94.0} & 29.0 & 36.0 & 27.0 & \textbf{7.0} & 22.0 & \textbf{68.0} & 0.0 & 75.0 & \textbf{69.0} & \textbf{32.0} \\ \midrule
\multicolumn{17}{c}{\textbf{Ablation Study}} \\ \midrule
\textbf{SEER (query-based)} & 79.0 & 39.0 & 97.0 & 39.2 & 80.0 & 61.0 & 95.0 & 24.0 & 33.0 & 26.0 & 4.0 & 24.0 & 69.0 & 0.0 & 83.0 & 67.5 & 30.5 \\ 
\textbf{SEER (w/o s2)} & 83.0 & 23.0 & 96.0 & 38.5 & 58.0 & 35.0 & 92.0 & 34.0 & 36.0 & 27.0 & 7.0 & 20.0 & 62.0 & 1.0 & 80.0 & 64.3 & 25.5 \\
\textbf{SEER (w/o s3)} & 89.0 & 27.0 & 64.0 & 41.5 & 91.0 & 60.0 & 95.0 & 24.0 & 39.0 & 35.0 & 6.0 & 23.0 & 65.0 & 0.0 & 0.0 & 56.1 & 30.1 \\
\bottomrule
\end{tabular}}
\end{table*}

% SEER(query-based) 79	39	97	39.23	80	61	95	24	33	26	4	24	69	0	83
% SEER(w/o s2) 83	23	96	38.46153846	58	35	92	34	36	27	7	20	62	1	80
% SEER(w/o s3) 89	27	64	41.53846154	91	60	95	24	39	35	6	23	65	0	0